\documentclass{article}
\usepackage[T1]{fontenc}
\usepackage{spconf}
\usepackage{multirow}
\usepackage{makecell}
\usepackage{amsmath}
\usepackage{amssymb}
\usepackage{booktabs}
\usepackage{graphicx}
\usepackage[
    colorlinks=true,
    urlcolor=black,
    linkcolor=black,
    citecolor=black
]{hyperref}

\title{ResCue: Residual Spatial Cueing for Language-Conditioned Imitation Learning}

\name{\shortstack{
Yanzhe Tang$^{1,2,\dagger}$, Xinyu Shao$^{1,2,\dagger}$, Yuxuan Hu$^{3}$, Siyu Chen$^{1}$, Chenxi Zhai$^{1}$ \\
Xi Cheng$^{1}$, Bowen Yang$^{2}$, Yajun Gao$^{2}$, Tongtong Cao$^{2}$, Xiu Li$^{1}$, Long Zeng$^{1,*}$
}\thanks{$^{\dagger}$Equal contribution; $^{*}$Corresponding author}}
\address{
$^{1}$ Tsinghua University, Shenzhen International Graduate School, Shenzhen, China\\
$^{2}$Huawei Technologies Co., Ltd., Shenzhen, China \\
$^{3}$National University of Singapore, Singapore
}

\begin{document}
\ninept
\maketitle

\begin{abstract}
Visual imitation policies must learn which image regions matter for action, whether applying the same operation to different objects or selecting a language-specified target. We introduce \textbf{ResCue} (\textbf{Res}idual Spatial \textbf{Cue}ing), which converts language-derived target information into a spatial condition for visual policy learning. Pretrained grounding produces a target mask; an independent cue encoder adds its features to spatial RGB features before further visual processing and pooling. This retains scene context while making the target explicit. With Diffusion Policy (DP), ResCue achieves 56.5\% average success across 15 simulated tasks, versus 33.1\% for DP and 40.7\% for the strongest baseline, $\pi_0$, with gains in both ordinary manipulation and instruction-dependent object selection. On three real-world bimanual tasks, it reaches 60.0\%, compared with 31.7\% for $\pi_0$. Matched-mask ablations favor residual fusion over input-level and pooled alternatives, while test-time interventions show that target-correct guidance matters for execution.
\end{abstract}


\section{INTRODUCTION}

Learning manipulation from RGB demonstrations couples two problems: identifying the visual features relevant to the target and mapping them to actions. When the same operation is demonstrated on different objects, the objective stays fixed while target appearance and geometry vary. With multiple candidates, the policy must additionally ground an instruction to a particular object. Vision-language-action (VLA) models jointly process images, language, and actions~\cite{zitkovich2023rt,kim2024openvla,black2026pi0}, but reliable target grounding remains challenging~\cite{huang2025roboground,yin2025codediffuser}. In both settings, the policy needs a consistent indication of the target while retaining the visual details required for manipulation. This motivates a common question: how should explicit target information condition the visual representation learned for control?

\begin{figure}[t]
\centering
\includegraphics[width=\columnwidth]{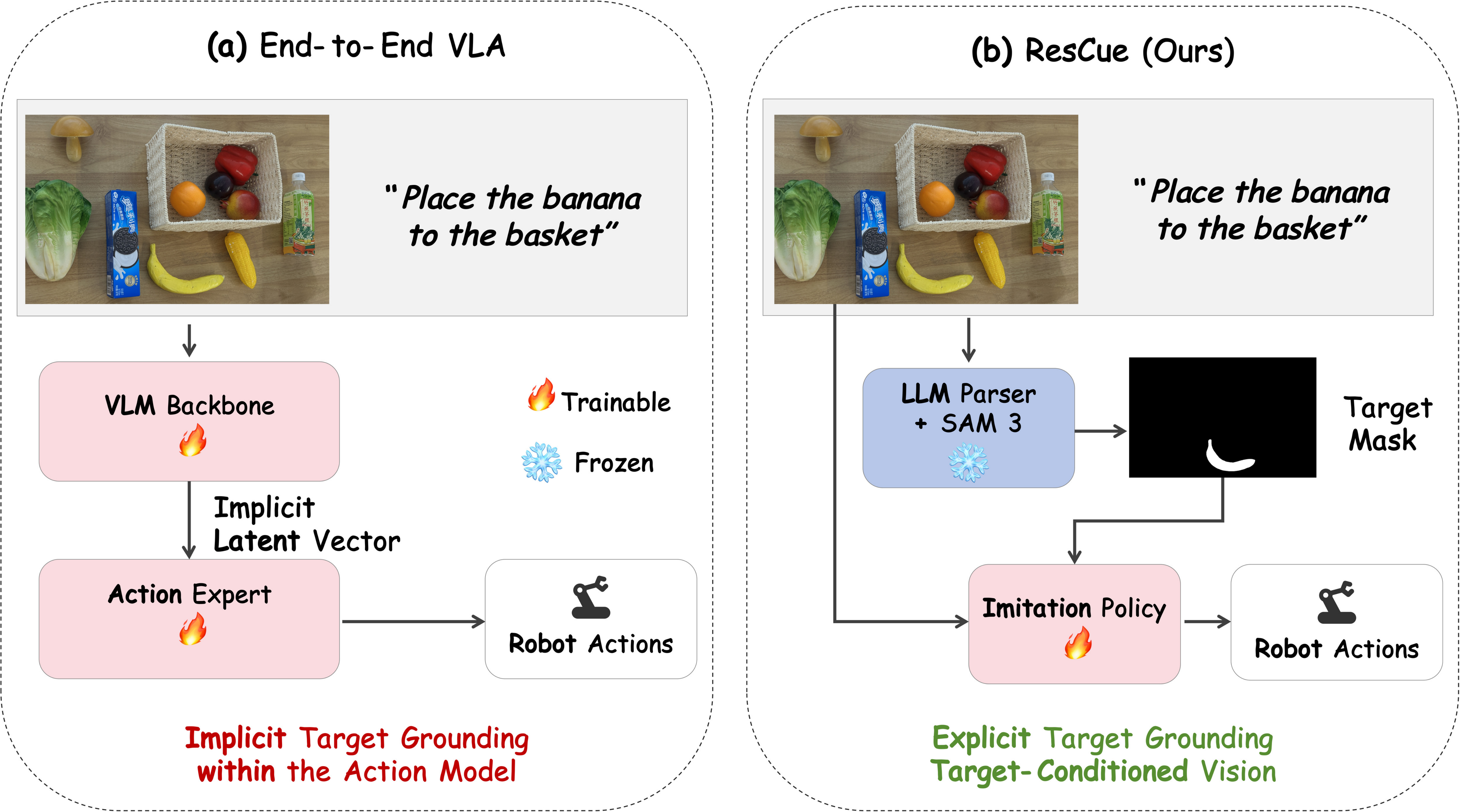}
\caption{\textbf{Explicit target grounding for visual policy learning.} (a) An end-to-end VLA pipeline processes the RGB observation and instruction through a VLM backbone and an action expert. (b) ResCue uses frozen language parsing and segmentation to obtain an explicit target mask. The mask and RGB observation jointly condition a trainable imitation policy, providing spatial target information for action generation.}
\label{fig:teaser}
\end{figure}

\begin{figure*}[!t]
\centering
\includegraphics[width=\textwidth]{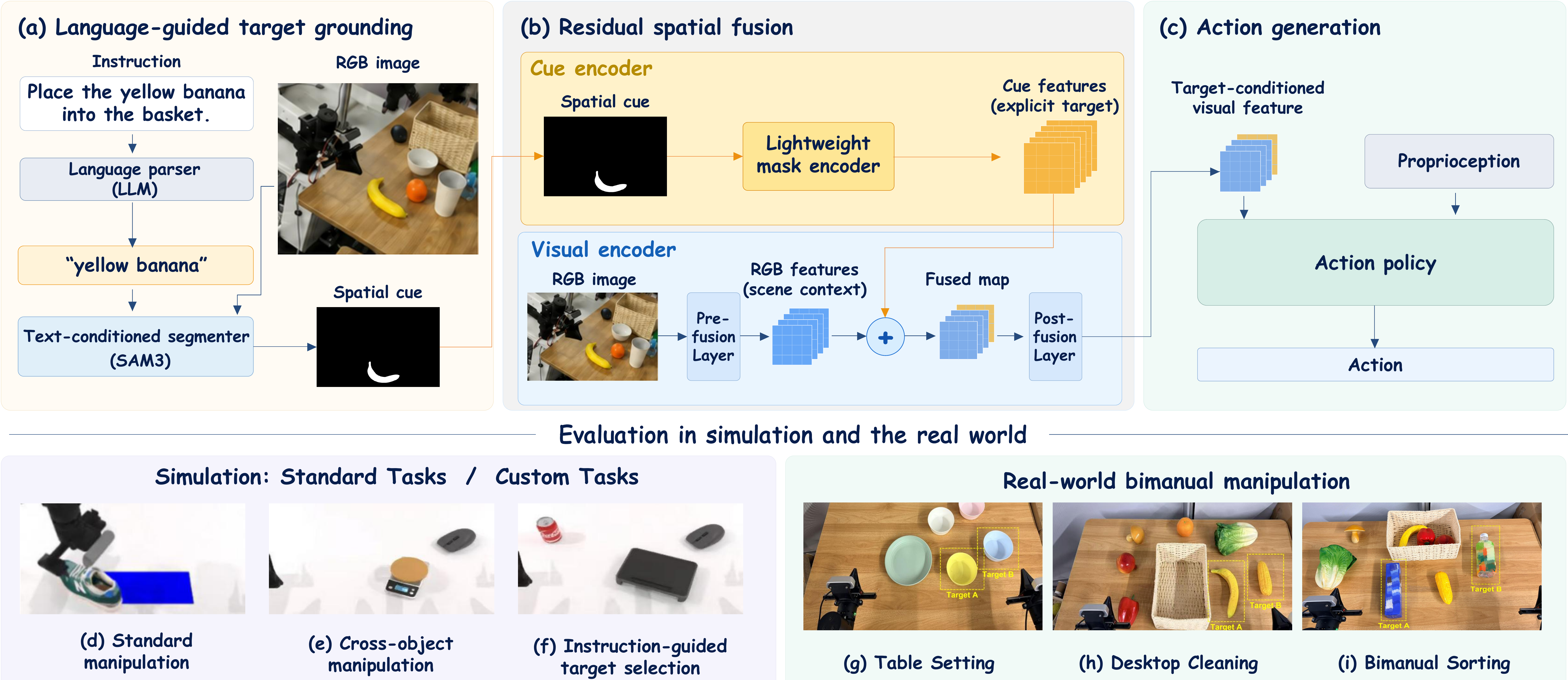}
\caption{\textbf{ResCue pipeline and evaluation tasks.} (a) A language parser extracts a target description, and a text-conditioned segmenter predicts its image-space mask. (b) An independent mask encoder provides explicit target features, while the RGB stream preserves scene context and object-specific visual information. Residual fusion combines the two streams before subsequent visual processing, yielding a target-conditioned representation. (c) This representation and proprioception condition the action policy to generate robot actions. (d)--(f) Simulation examples cover standard manipulation, cross-object manipulation, and instruction-guided target selection. (g)--(i) Real-world bimanual tasks comprise Table Setting, Desktop Cleaning, and Bimanual Sorting.}
\label{fig:method_overview}
\end{figure*}

\textbf{Spatial intermediates for manipulation.} Explicit spatial representations connect language reasoning with robot execution. MOKA uses marked images to elicit point-based manipulation affordances from vision-language models~\cite{liu2024moka}, while ReKep expresses manipulation objectives through relational keypoint constraints~\cite{huang2024rekep}. RoboMAP predicts continuous, adaptive affordance heatmaps that represent spatial uncertainty, including valid regions for object-free placement~\cite{shao2025robomap}. These approaches motivate spatial intermediates beyond global language embeddings. Our focus is complementary to improving grounding representations: we study how a predicted target cue conditions the visual features learned for continuous imitation control.

\textbf{Spatial cue integration for policy learning.} Existing approaches integrate spatial cues into policies in different ways. RoboGround incorporates object and placement masks through input-channel conditioning and mask-guided patch attention~\cite{huang2025roboground}. CodeDiffuser uses VLM-generated code to construct 3D attention maps for a diffusion policy~\cite{yin2025codediffuser}. Image-space alternatives include AimBot's end-effector-related visual overlays~\cite{dai2025aimbot} and VP-VLA's target and placement prompts with auxiliary grounding supervision~\cite{wang2026vpvla}. These methods differ both in the spatial information they provide and in where that information interacts with the policy. Beyond obtaining a spatial cue, the policy must integrate it with RGB features so that target information can guide visual processing while retaining scene context.

To address this question, we introduce \textbf{ResCue} (\textbf{Res}idual Spatial \textbf{Cue}ing), which converts a language description into a target mask and injects its encoded features into the policy's RGB stream through residual addition (Fig.~\ref{fig:teaser}). An independent mask encoder produces spatial cue features that are fused with spatial RGB features before subsequent visual processing and pooling. This allows the later visual layers to form a target-conditioned representation while retaining the RGB scene context needed for object-specific manipulation. Across objects, the mask provides a shared conditioning format---act on the indicated region---while the RGB stream preserves object-specific appearance and geometry. The same interface can highlight the task-relevant object in standard manipulation and disambiguate the intended object when multiple candidates are present.

Our contributions are twofold: (1) a residual spatial-cue interface that independently encodes a target mask and fuses it with spatial RGB features before subsequent visual processing and pooling; and (2) a controlled evaluation with Diffusion Policy (DP)~\cite{chi2023diffusion} on 15 simulated and three real-world tasks, spanning standard manipulation, cross-object manipulation, and instruction-dependent target selection. Matched-mask comparisons provide a controlled evaluation of fusion interfaces, while test-time cue interventions examine whether and how the learned policy uses target information.

\section{METHODOLOGY}
\label{sec:methodology}

ResCue grounds the target in image coordinates and uses that cue to condition visual processing before action generation (Fig.~\ref{fig:method_overview}).

\subsection{Grounding Language into a Spatial Cue}
\label{sec:spatial_cue}

\textbf{Problem setting.} A task $\tau$ defines the manipulation objective, and an instruction $l$ specifies the target object. We train a separate policy for each task. ResCue extracts a mask for the task-relevant manipulated object, particularly the object whose identity varies across demonstrations or is specified by language. At time $t$, the inputs are an RGB image $I_t\in\mathbb{R}^{H\times W\times3}$ and proprioception $q_t\in\mathbb{R}^{d_q}$.

An LLM-based parser $\mathcal{P}$ extracts a target description $c$, which conditions the segmenter $\mathcal{S}$ (SAM~3~\cite{carion2025sam}):
\begin{equation}
  c=\mathcal{P}(l),\qquad M_t=\mathcal{S}(I_t,c).
  \label{eq:grounding}
\end{equation}
The binary mask $M_t\in\{0,1\}^{H\times W}$ encodes the target's location and silhouette. It provides a common target-indication format across objects, while RGB retains object-specific appearance, geometry, and placement context.

\subsection{Residual Spatial Conditioning}
\label{sec:residual_injection}

Let $\Phi_{\mathrm{pre}}$ and $\Phi_{\mathrm{post}}$ denote the visual encoder before and after fusion, and $\Phi_{\mathrm{cue}}$ a mask-only convolutional encoder. We compute
\begin{equation}
\begin{aligned}
  F_{\mathrm{rgb},t} &= \Phi_{\mathrm{pre}}(I_t), \\
  F_{\mathrm{cue},t} &= \Phi_{\mathrm{cue}}(M_t), \\
  \widetilde F_t &= F_{\mathrm{rgb},t}+F_{\mathrm{cue},t}, \\
  v_t &= \operatorname{Pool}\!\left(\Phi_{\mathrm{post}}(\widetilde F_t)\right).
\end{aligned}
\label{eq:residual_fusion}
\end{equation}
Both feature maps have shape $C\times h\times w$, enabling location-wise residual addition. The fused map is processed by the subsequent visual layers before pooling, allowing target cues to condition spatial feature extraction together with RGB scene information.

In our DP implementation, a three-layer convolutional cue encoder produces features matching the RGB encoder's stem output. Residual addition is applied after the stem and before the residual stages.

\subsection{Policy Learning and Execution}
\label{sec:action_generation}

Visual features and proprioception form $z_t=\operatorname{Concat}(v_t,q_t)$. A history of $K$ observations conditions action generation:
\begin{equation}
  Z_t=(z_{t-K+1},\ldots,z_t),\qquad A_t\sim\pi_{\phi_\tau}(\cdot\mid Z_t).
\label{eq:action_conditioning}
\end{equation}
The action chunk $A_t$ contains continuous commands and gripper states for both arms.

We train DP's conditional 1D U-Net using its standard noise-prediction objective~\cite{chi2023diffusion}. The resulting action loss jointly optimizes the action network, RGB encoder, and cue encoder, while the grounding models remain frozen. Predicted masks are used during both training and evaluation.

At execution, the parser runs once per instruction, and SAM~3 predicts a mask for each new observation. RGB and mask are resized consistently; real-world segmentation precedes resizing of the high-resolution image. Deployment is configured for 10~Hz policy updates and a 300~Hz low-level controller, with action interpolation.

\section{EXPERIMENTS}
\label{sec:experiments}

We evaluate ResCue in both simulation and real-world experiments on a bimanual platform. Simulation experiments combine standard manipulation benchmarks with custom tasks examining manipulation across different objects and instruction-guided target selection. Real-world experiments evaluate target-directed manipulation in cluttered scenes. We then compare mask fusion strategies and analyze the effects of test-time target masks.

\begin{table}[!t]
\centering
\caption{Simulation success rates (\%) over 100 episodes per task. Custom tasks evaluate cross-object manipulation and instruction-guided target selection. In all tables, \textbf{bold} and \underline{underline} mark the best and second-best values.}
\label{tab:performance_comparison}
\vspace{3pt}
\footnotesize
\renewcommand{\arraystretch}{1.12}
\setlength{\tabcolsep}{2.3pt}
\begin{tabular*}{\columnwidth}{@{\extracolsep{\fill}}lccccc@{}}
\toprule
\multirow{2}{*}{\textbf{Task}} & \multicolumn{5}{c}{\textbf{Success Rate (\%)}} \\
\cmidrule(lr){2-6}
& \textbf{DP} & \textbf{ACT} & \textbf{RDT} & \textbf{$\pi_0$} & \textbf{ResCue} \\
\midrule
\multicolumn{6}{c}{\textit{Standard RoboTwin 2.0 Tasks}} \\
\midrule
Adjust Bottle & \textbf{97.0} & \textbf{97.0} & 81.0 & 90.0 & \underline{96.0} \\
Click Bell & 54.0 & 58.0 & \underline{80.0} & 44.0 & \textbf{83.0} \\
Hanging Mug & 8.0 & 7.0 & \underline{23.0} & 11.0 & \textbf{24.0} \\
Lift Pot & 39.0 & \underline{88.0} & 72.0 & 84.0 & \textbf{90.0} \\
Place Can Basket & 18.0 & 1.0 & 19.0 & \underline{41.0} & \textbf{42.0} \\
Place Cans Plasticbox & \underline{40.0} & 16.0 & 6.0 & 34.0 & \textbf{88.0} \\
Place Empty Cup & 37.0 & \textbf{61.0} & \underline{56.0} & 37.0 & 49.0 \\
Place Shoe & 23.0 & 5.0 & \underline{35.0} & 28.0 & \textbf{45.0} \\
Shake Bottle & 65.0 & 74.0 & 74.0 & \textbf{97.0} & \underline{93.0} \\
\midrule
\textbf{Average} & 42.3 & 45.2 & 49.6 & \underline{51.8} & \textbf{67.8} \\
\midrule
\multicolumn{6}{c}{\textit{Custom Manipulation Tasks}} \\
\midrule
\multicolumn{6}{l}{\textit{Cross-object manipulation}} \\
Move Single Object to Basket & 27.0 & 4.0 & 27.0 & \underline{42.0} & \textbf{67.0} \\
Move Single Object to Stand & \underline{32.0} & 0.0 & 19.0 & 23.0 & \textbf{54.0} \\
Move Single Object to Scale & 19.0 & 0.0 & 7.0 & \underline{27.0} & \textbf{38.0} \\
\multicolumn{6}{l}{\textit{Instruction-guided target selection}} \\
Pick Specific Object to Basket & 9.0 & 5.0 & \underline{16.0} & 7.0 & \textbf{25.0} \\
Pick Specific Object to Stand & 17.0 & 1.0 & 8.0 & \underline{19.0} & \textbf{34.0} \\
Pick Specific Object to Scale & 12.0 & 7.0 & 7.0 & \textbf{26.0} & \underline{19.0} \\
\midrule
\textbf{Average} & 19.3 & 2.8 & 14.0 & \underline{24.0} & \textbf{39.5} \\
\midrule
\textbf{All Tasks Average} & 33.1 & 28.3 & 35.3 & \underline{40.7} & \textbf{56.5} \\
\bottomrule
\end{tabular*}
\end{table}

\subsection{Experimental Setup}

\textbf{Simulation tasks.} We evaluate nine standard RoboTwin 2.0 tasks~\cite{chen2025robotwin} and six custom tasks. The custom tasks comprise two groups of three tasks each.

The three \emph{Move Single Object} tasks require moving an object to a task-specific destination: a basket, a stand, or a scale. Demonstrations cover different objects, and each training and evaluation episode contains one manipulation object. These tasks examine learning a shared manipulation operation across object variations.

The three \emph{Pick Specific Object} tasks present two candidate objects and require moving the language-specified target to the designated destination. These tasks examine instruction following during target-directed manipulation. Custom-task evaluation uses the training objects in varied scene configurations.

\textbf{Real-world tasks.} We evaluate \emph{Table Setting}, \emph{Desktop Cleaning}, and \emph{Bimanual Sorting} on an Aloha-AgileX bimanual robot. These tasks involve target selection based on color, semantic category, and spatial arrangement, respectively.

\textbf{Baselines.} We compare ResCue with language-agnostic DP~\cite{chi2023diffusion} and ACT~\cite{zhao2023learning}, and pretrained VLA baselines RDT~\cite{liu2025rdt} and $\pi_0$~\cite{black2026pi0}. RDT and $\pi_0$ process language directly, while ResCue converts target language into a spatial cue on a DP backbone.

\textbf{Protocol.} We use 50 demonstrations per standard simulation task and 100 per custom simulation or real-world task. All baselines are fine-tuned on the same task data using their official recipes. We report success over 100 evaluation episodes per simulated task and 20 trials per real-world task. Success requires completing the specified operation on the intended object, and averages weight tasks equally.

\subsection{Simulation Results}
\label{sec:simulation_results}

Across all 15 tasks, ResCue reaches 56.5\% success, exceeding DP (33.1\%), RDT (35.3\%), and $\pi_0$ (40.7\%) (Table~\ref{tab:performance_comparison}). ResCue also outperforms both pretrained VLA baselines, RDT and $\pi_0$. We examine standard manipulation, cross-object manipulation, and instruction-guided target selection.

\textbf{Standard benchmark.} On standard tasks, ResCue achieves 67.8\%, compared with 42.3\% for DP and 51.8\% for $\pi_0$. These results show the effectiveness of explicit target conditioning across the benchmark manipulation tasks.

\textbf{Cross-object manipulation.} On the three single-object custom tasks, ResCue reaches 53.0\%, versus 26.0\% for DP and 30.7\% for $\pi_0$. Despite variations in object appearance, shape, and position, the mask consistently identifies the manipulated object. The policy receives this target-region information together with RGB features containing object-specific details and destination context, providing a consistent indication of the object's task role across demonstrations.

\textbf{Instruction-guided target selection.} Across the three two-candidate tasks, ResCue achieves an average success rate of 26.0\%, compared with 10.3\% for RDT and 17.3\% for $\pi_0$. This comparison suggests that converting language into an explicit visual cue for the visuomotor policy (Fig.~\ref{fig:teaser}(b)) is more effective than relying on VLA models to infer the target implicitly from image-language inputs (Fig.~\ref{fig:teaser}(a)).

\subsection{Real-World Bimanual Manipulation}
\label{sec:real_world}

\begin{table}[!b]
\centering
\caption{Real-world success rates (\%) on the Aloha-AgileX, using 100 in-domain demonstrations and 20 evaluation trials per task.}
\label{tab:real_world_results}
\vspace{3pt}
\footnotesize
\renewcommand{\arraystretch}{1.12}
\setlength{\tabcolsep}{3pt}
\begin{tabular*}{\columnwidth}{@{\extracolsep{\fill}}lccc@{}}
\toprule
\textbf{Task} & \textbf{DP} & \textbf{$\pi_0$} & \textbf{ResCue} \\
\midrule
Table Setting & 20 & \underline{30} & \textbf{75} \\
Desktop Cleaning & \underline{35} & 20 & \textbf{50} \\
Bimanual Sorting & 30 & \underline{45} & \textbf{55} \\
\midrule
\textbf{Average} & 28.3 & \underline{31.7} & \textbf{60.0} \\
\bottomrule
\end{tabular*}
\end{table}

Three cluttered tasks on an Aloha-AgileX bimanual robot test distinct aspects of target-directed manipulation. \emph{Table Setting} requires placing the instructed bowl onto a plate among bowls with identical geometry but different colors. \emph{Desktop Cleaning} tests category-based selection among similarly colored objects, such as yellow corn and a yellow banana. \emph{Bimanual Sorting} requires selecting the appropriate arm according to the target's location and each arm's physical reach. Evaluation uses new target poses and unseen distractors. Success requires selecting the target, grasping it stably, and placing it with the appropriate arm.

ResCue leads on all three tasks, averaging 60.0\% against 28.3\% for DP and 31.7\% for $\pi_0$ (Table~\ref{tab:real_world_results}). These results demonstrate target-directed bimanual manipulation under color, category, and spatial requirements.

\subsection{Does the Cue Interface Matter?}
\label{sec:fusion_ablation}

\begin{table}[!t]
\centering
\caption{Fusion ablation: average success rates (\%) with 100 evaluation episodes per task. Each column is a separately trained policy; mask-conditioned variants use the same target masks.}
\label{tab:ablation_fusion}
\vspace{3pt}
\footnotesize
\renewcommand{\arraystretch}{1.12}
\setlength{\tabcolsep}{1.2pt}
\begin{tabular*}{\columnwidth}{@{\extracolsep{\fill}}lccccc@{}}
\toprule
\textbf{Task Split}
& \makecell{\textbf{Unprompted}\\\textbf{DP}}
& \makecell{\textbf{Pixel}\\\textbf{Overlay}}
& \makecell{\textbf{Early}\\\textbf{Concat}}
& \makecell{\textbf{Two-Stream}\\\textbf{Late Fusion}}
& \textbf{ResCue} \\
\midrule
Standard & 42.3 & \underline{50.1} & 43.4 & 44.4 & \textbf{67.8} \\
Custom & 19.3 & 20.0 & \underline{22.2} & 19.3 & \textbf{39.5} \\
\midrule
\textbf{All Tasks} & 33.1 & \underline{38.1} & 34.9 & 34.4 & \textbf{56.5} \\
\bottomrule
\end{tabular*}
\end{table}

We train four mask-conditioned DP variants using the same target masks: \emph{pixel overlay} highlights the target in RGB; \emph{early concatenation} adds an input mask channel; \emph{two-stream late fusion} concatenates independently encoded and pooled RGB and mask features; and \emph{residual spatial conditioning} adds cue features before subsequent visual processing.
All variants share training and evaluation data and optimization settings, enabling a controlled comparison of cue-integration interfaces.

ResCue reaches 56.5\% overall success, versus 38.1\% for overlay, 34.9\% for early concatenation, and 34.4\% for late fusion (Table~\ref{tab:ablation_fusion}). It leads on both standard and custom tasks, with an overall margin of 18.4 percentage points over the strongest alternative.

Early concatenation processes RGB and mask information jointly from the input while preserving spatial alignment. Late fusion combines the two streams after pooling, when the fusion inputs no longer retain an explicit two-dimensional feature grid. ResCue instead merges independently encoded RGB and mask features at corresponding spatial locations. The subsequent visual layers process these fused features before pooling, making target information available during spatial feature extraction. The results favor this combination of independent encoding and pre-pooling spatial fusion.

\subsection{How Does the Policy Use the Cue?}
\label{sec:cue_validity}

We hold each trained ResCue policy fixed and change only its test-time mask; RGB and proprioception update normally. Table~\ref{tab:prompt_validity} compares \emph{No Mask} (all zeros), \emph{Wrong Object Mask} (the mask of a random non-target scene object, such as a receptacle), \emph{First-Frame Mask} (reuse $M_0$), \emph{Perturbed Mask} (random erosion or dilation and spatial shifts of updated masks), and the default \emph{Per-Step Mask} (a fresh prediction per observation).

\textbf{Cue availability.} Removing the mask reduces overall success from 56.5\% to 46.9\%, a drop of 9.6 percentage points. The trained policy therefore benefits from target information during execution.

\textbf{Target correctness.} Wrong-object masks reduce success to 24.6\%, 22.3 percentage points below mask removal. Under these conditions, an incorrect spatial target degrades task completion more strongly than omitting the cue.

\textbf{Cue updates.} A first-frame mask keeps the target region fixed while RGB observations continue to change; per-step prediction updates the spatial cue using the current observation. This raises standard-task success from 55.7\% to 67.8\%, a gain of 12.1 percentage points. The custom-task average changes from 39.2\% to 39.5\%, a gain of 0.3 points. The benefit of refreshing target localization is therefore concentrated in the standard tasks.

\textbf{Spatial precision.} Under the evaluated erosion, dilation, and spatial-shift perturbations, ResCue retains 50.5\% overall success, 6.0 percentage points below the default and 3.6 points above mask removal. These perturbations reduce performance while retaining part of the benefit of target guidance.

\begin{table}[!t]
\centering
\caption{Test-time mask interventions: average success rates (\%) over 100 episodes per task and condition, without retraining. Per-Step Mask is the default; wrong cues indicate non-target scene objects.}
\label{tab:prompt_validity}
\vspace{3pt}
\footnotesize
\renewcommand{\arraystretch}{1.12}
\setlength{\tabcolsep}{1.1pt}
\begin{tabular*}{\columnwidth}{@{\extracolsep{\fill}}lccccc@{}}
\toprule
\textbf{Task Split}
& \makecell{\textbf{No}\\\textbf{Mask}}
& \makecell{\textbf{Wrong}\\\textbf{Object}\\\textbf{Mask}}
& \makecell{\textbf{First-Frame}\\\textbf{Mask}}
& \makecell{\textbf{Perturbed}\\\textbf{Mask}}
& \makecell{\textbf{Per-Step}\\\textbf{Mask}} \\
\midrule
Standard & 57.2 & 23.1 & 55.7 & \underline{60.2} & \textbf{67.8} \\
Custom & 31.5 & 26.8 & \underline{39.2} & 35.8 & \textbf{39.5} \\
\midrule
\textbf{All Tasks} & 46.9 & 24.6 & 49.1 & \underline{50.5} & \textbf{56.5} \\
\bottomrule
\end{tabular*}
\end{table}

\section{CONCLUSION AND LIMITATIONS}
\label{sec:conclusion}

ResCue makes the manipulation target explicit within the visual representation used for action learning through independent cue encoding and residual spatial fusion. Our DP instantiation improves average success across simulated manipulation tasks and real-world bimanual tasks, covering shared operations across objects and instruction-dependent target selection. Matched-mask comparisons favor this spatial interface over input-level and pooled alternatives, while test-time mask interventions show that target-correct guidance matters for execution. Together, these results support target-conditioned vision as an effective interface connecting pretrained grounding with action learning.

\textbf{Limitations.} Our experiments instantiate ResCue only with DP. Although residual spatial cueing is not specific to diffusion-based action generation and may be applicable to other visual policies, its effectiveness across policy architectures remains to be validated. Second, ResCue depends on the quality of upstream grounding: in real-world deployment, we observed unstable masks for transparent, highly reflective, or heavily occluded objects, while our test-time interventions further show that incorrect target masks can propagate directly to policy execution and substantially reduce task success. Finally, SAM~3 segmentation introduced inference latency that limited rapid responses in dynamic scenes. For fixed tasks and deployment settings, training a task-specific lightweight segmentation model may improve grounding reliability while reducing inference overhead and hardware requirements.

\clearpage
\bibliographystyle{IEEEbib}
\bibliography{refs}

@inproceedings{chen2025robotwin,
  title={{RoboTwin} 2.0: A Scalable Data Generator and Benchmark with Strong Domain Randomization for Robust Bimanual Robotic Manipulation},
  author={Chen, Tianxing and Chen, Zanxin and Chen, Baijun and Cai, Zijian and Liu, Yibin and Li, Zixuan and Liang, Qiwei and Lin, Xianliang and Ge, Yiheng and Gu, Zhenyu and others},
  booktitle={Proceedings of the International Conference on Machine Learning (ICML)},
  year={2026}
}

@inproceedings{carion2025sam,
  title={{SAM} 3: Segment Anything with Concepts},
  author={Carion, Nicolas and Gustafson, Laura and Hu, Yuan-Ting and Debnath, Shoubhik and Hu, Ronghang and Suris Coll-Vinent, Didac and Ryali, Chaitanya and Alwala, Kalyan Vasudev and Khedr, Haitham and Huang, Andrew and others},
  booktitle={Proceedings of the International Conference on Learning Representations (ICLR)},
  year={2026}
}

@inproceedings{chi2023diffusion,
  title={Diffusion Policy: Visuomotor Policy Learning via Action Diffusion},
  author={Chi, Cheng and Feng, Siyuan and Du, Yilun and Xu, Zhenjia and Cousineau, Eric and Burchfiel, Benjamin C. M. and Song, Shuran},
  booktitle={Proceedings of Robotics: Science and Systems (RSS)},
  year={2023},
  doi={10.15607/RSS.2023.XIX.026}
}

@inproceedings{zhao2023learning,
  title={Learning Fine-Grained Bimanual Manipulation with Low-Cost Hardware},
  author={Zhao, Tony Z. and Kumar, Vikash and Levine, Sergey and Finn, Chelsea},
  booktitle={Proceedings of Robotics: Science and Systems (RSS)},
  year={2023},
  doi={10.15607/RSS.2023.XIX.016}
}

@inproceedings{liu2025rdt,
  title={{RDT}: A Diffusion Foundation Model for Bimanual Manipulation},
  author={Liu, Songming and Wu, Lingxuan and Li, Bangguo and Tan, Hengkai and Chen, Huayu and Wang, Zhengyi and Xu, Ke and Su, Hang and Zhu, Jun},
  booktitle={Proceedings of the International Conference on Learning Representations (ICLR)},
  year={2025}
}

@inproceedings{black2026pi0,
  title={{$\pi_0$}: A Vision-Language-Action Flow Model for General Robot Control},
  author={Black, Kevin and Brown, Noah and Driess, Danny and Esmail, Adnan and Equi, Michael Robert and Finn, Chelsea and Fusai, Niccolo and Groom, Lachy and Hausman, Karol and Ichter, Brian and others},
  booktitle={Proceedings of Robotics: Science and Systems (RSS)},
  year={2025},
  doi={10.15607/RSS.2025.XXI.010}
}

@inproceedings{dai2025aimbot,
  title={{AimBot}: A Simple Auxiliary Visual Cue to Enhance Spatial Awareness of Visuomotor Policies},
  author={Dai, Yinpei and Lee, Jayjun and Zhang, Yichi and Ma, Ziqiao and Yang, Jianing and Zadeh, Amir and Li, Chuan and Fazeli, Nima and Chai, Joyce},
  booktitle={Proceedings of the Conference on Robot Learning (CoRL)},
  volume={305},
  series={Proceedings of Machine Learning Research},
  pages={2409--2429},
  year={2025}
}

@inproceedings{kim2024openvla,
  title={{OpenVLA}: An Open-Source Vision-Language-Action Model},
  author={Kim, Moo Jin and Pertsch, Karl and Karamcheti, Siddharth and Xiao, Ted and Balakrishna, Ashwin and Nair, Suraj and Rafailov, Rafael and Foster, Ethan P. and Sanketi, Pannag R. and Vuong, Quan and others},
  booktitle={Proceedings of the Conference on Robot Learning (CoRL)},
  volume={270},
  series={Proceedings of Machine Learning Research},
  pages={2679--2713},
  year={2025}
}

@inproceedings{zitkovich2023rt,
  title={{RT}-2: Vision-Language-Action Models Transfer Web Knowledge to Robotic Control},
  author={Zitkovich, Brianna and Yu, Tianhe and Xu, Sichun and Xu, Peng and Xiao, Ted and Xia, Fei and Wu, Jialin and Wohlhart, Paul and Welker, Stefan and Wahid, Ayzaan and others},
  booktitle={Proceedings of the Conference on Robot Learning (CoRL)},
  volume={229},
  series={Proceedings of Machine Learning Research},
  pages={2165--2183},
  year={2023}
}

@inproceedings{liu2024moka,
  title={{MOKA}: Open-World Robotic Manipulation through Mark-Based Visual Prompting},
  author={Fang, Kuan and Liu, Fangchen and Abbeel, Pieter and Levine, Sergey},
  booktitle={Proceedings of Robotics: Science and Systems (RSS)},
  year={2024}
}

@inproceedings{huang2024rekep,
  title={{ReKep}: Spatio-Temporal Reasoning of Relational Keypoint Constraints for Robotic Manipulation},
  author={Huang, Wenlong and Wang, Chen and Li, Yunzhu and Zhang, Ruohan and Fei-Fei, Li},
  booktitle={Proceedings of the Conference on Robot Learning (CoRL)},
  volume={270},
  series={Proceedings of Machine Learning Research},
  pages={4573--4602},
  year={2025}
}

@article{shao2025robomap,
  title={More than A Point: Capturing Uncertainty with Adaptive Affordance Heatmaps for Spatial Grounding in Robotic Tasks},
  author={Shao, Xinyu and Tang, Yanzhe and Xie, Pengwei and Zhou, Kaiwen and Zhuang, Yuzheng and Quan, Xingyue and Hao, Jianye and Zeng, Long and Li, Xiu},
  journal={arXiv preprint arXiv:2510.10912},
  year={2025}
}

@inproceedings{huang2025roboground,
  title={{RoboGround}: Robotic Manipulation with Grounded Vision-Language Priors},
  author={Huang, Haifeng and Chen, Xinyi and Chen, Yilun and Li, Hao and Han, Xiaoshen and Wang, Zehan and Wang, Tai and Pang, Jiangmiao and Zhao, Zhou},
  booktitle={Proceedings of the IEEE/CVF Conference on Computer Vision and Pattern Recognition (CVPR)},
  pages={22540--22550},
  year={2025}
}

@inproceedings{yin2025codediffuser,
  title={{CodeDiffuser}: Attention-Enhanced Diffusion Policy via {VLM}-Generated Code for Instruction Ambiguity},
  author={Yin, Guang and Li, Yitong and Wang, Yixuan and McConachie, Dale and Shah, Paarth and Hashimoto, Kunimatsu and Zhang, Huan and Liu, Katherine and Li, Yunzhu},
  booktitle={Proceedings of Robotics: Science and Systems (RSS)},
  year={2025},
  doi={10.15607/RSS.2025.XXI.072}
}

@article{wang2026vpvla,
  title={{VP-VLA}: Visual Prompting as an Interface for Vision-Language-Action Models},
  author={Wang, Zixuan and Chen, Yuxin and Liu, Yuqi and Ye, Jinhui and Chen, Pengguang and Lu, Changsheng and Liu, Shu and Yu, Bei and Jia, Jiaya},
  journal={arXiv preprint arXiv:2603.22003},
  year={2026}
}

\end{document}